\pdfoutput=1

\documentclass[11pt]{article}

\usepackage[preprint]{acl}

\PassOptionsToPackage{dvipsnames}{xcolor}
\usepackage{xcolor}         

\usepackage{times}
\usepackage{latexsym}

\usepackage[T1]{fontenc}

\usepackage[utf8]{inputenc}

\usepackage{microtype}

\usepackage{inconsolata}

\usepackage{graphicx}

\usepackage[T1]{fontenc}    
\usepackage{hyperref}       
\usepackage{url}            
\usepackage{booktabs}       
\usepackage{amsfonts}       
\usepackage{nicefrac}       
\usepackage{pgfplots}
\usepackage{pgfplotstable}
\usepackage{sansmath}
\usepackage{helvet}
\usepackage{amsmath}
\usepackage{tcolorbox}
\usepackage{float}          
\usepackage{multirow}       
\usepackage{tabularx}
\usepackage{tikz}
\usetikzlibrary{arrows.meta, positioning}
\usepackage[normalem]{ulem}
\usepackage{makecell}

\pgfplotsset{compat=1.18}

\title{RAG-RL: Advancing Retrieval-Augmented Generation via RL and Curriculum Learning}

\author{
    \textbf{Jerry Huang}\textsuperscript{1}\thanks{This work was done while Jerry was a Research Intern at NewsBreak. Correspondence to: jerry8@illinois.edu and tozhang@illinois.edu.}, 
    \textbf{Siddarth Madala}\textsuperscript{1}, 
    \textbf{Risham Sidhu}\textsuperscript{1},
    \textbf{Cheng Niu}\textsuperscript{2}, \\
    \textbf{Hao Peng}\textsuperscript{1}, 
    \textbf{Julia Hockenmaier}\textsuperscript{1}, 
    \textbf{Tong Zhang}\textsuperscript{1} \\
    \textsuperscript{1}University of Illinois at Urbana-Champaign \\
    \textsuperscript{2}NewsBreak
}

\begin{document}
\maketitle

\begin{abstract}
Retrieval-augmented generation (RAG) systems rely on retrieval models for identifying relevant contexts and answer generation models for utilizing those contexts. However, retrievers exhibit imperfect recall and precision, limiting downstream performance. We introduce \mbox{\textbf{RAG-RL}}, an answer generation model trained not only to produce answers but also to identify and cite relevant information from larger sets of retrieved contexts, shifting some of the burden of identifying relevant documents from the retriever to the answer generator. Our approach uses \textit{curriculum learning}, where the model is first trained on easier examples that include only relevant contexts. Our experiments show that these training samples enable models to acquire citation and reasoning skills with greater sample efficiency and generalizability, demonstrating strong model performance even as the number of irrelevant passages increases. We benchmark our methods on three open-domain multi-hop question answering datasets and report significant gains in answer and citation accuracy. Our experiments provide empirical insights into how easier training samples can give models stronger signals for learning specific skills (e.g., citation generation) and how different components of post-training (e.g., training set construction, rule-based rewards, training sample ordering, etc.) impact final model performance.
\end{abstract}
\section{Introduction}

\begin{figure}[h]
  \centering
  \resizebox{0.4\textwidth}{!}{%
    \begin{tikzpicture}[
      node distance=0.5cm,
      every node/.style={draw, rounded corners, align=left, font=\small, text width=7.5cm, inner sep=6pt},
      question/.style={fill=blue!5},
      hop1/.style={fill=gray!10},
      hop2/.style={fill=gray!15},
      answer/.style={fill=green!10},
      arrow/.style={-{Latex[scale=1]}, thick}
    ]

    \node[question] (q) {\textbf{Question:} Who is the maternal grandfather of \uline{Floris De Voogd}?};

    \node[hop1, below=of q] (hop1) {\textbf{Hop 1:} \textbf{Title: Floris de Voogd} \\
    ``Floris de Voogd (c. 1228–1258) ... son of \uline{Floris IV, Count of Holland} and \uline{Matilda of Brabant} ...''};

    \node[hop2, below=of hop1] (hop2) {\textbf{Hop 2:} \textbf{Title: Matilda of Brabant, Countess of Holland} \\
    ``Matilda ... was the daughter of \uline{Henry I, Duke of Brabant} and Mathilde of Flanders ...''};

    \node[answer, below=of hop2] (a) {
      \textcolor{green!10!black}{
        \textbf{<reasoning>}\\
          The question asks for the maternal grandfather of Floris de Voogd. The passage \textit{"Floris de Voogd"} provides information about Floris de Voogd's parents: Floris IV, Count of Holland and Matilda of Brabant.\\
          ...\\
          Since Matilda of Brabant is the mother of Floris de Voogd, Henry I, Duke of Brabant is the maternal grandfather of Floris de Voogd.\\
          \textbf{</reasoning>}\\
          \textbf{<answer>}\\
          Final answer: Henry I, Duke of Brabant\\
          Supporting passages: Matilda of Brabant, Countess of Holland, Floris de Voogd\\
          \textbf{</answer>}
      }
    };

    \draw[arrow] (q) -- (hop1);
    \draw[arrow] (hop1) -- (hop2);
    \draw[arrow] (hop2) -- (a);

    \end{tikzpicture}
  }
  \caption{An example of a multi-hop reasoning chain taken from the MuSiQue dataset. RAG-RL generated the reasoning trace and final answer/citations observed in the green block.}
  \label{fig:multihop-floris}
\end{figure}
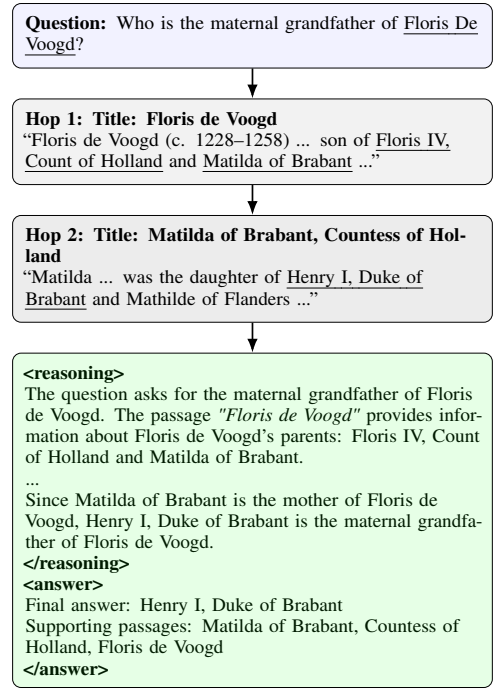

Retrieval-augmented generation (RAG;~\citealp{guu2020realm, lewis2020rag, wang2024rear}) relies on retrieval and generation models working together to retrieve and integrate external contexts effectively for answering questions or generating content. While previous works have made significant progress in improving these systems by optimizing retrieval and reranking models~\cite{zhang2024endtoendbeamretrievalmultihop, gutiérrez2025ragmemory, weller2025rank1}, challenges persist when it comes to retrieving relevant real-world contexts that require reasoning, especially those whose relevance goes beyond semantic similarity~\cite{su2024brightrealisticchallengingbenchmark}. Moreover, prior work has demonstrated that generative models often struggle to effectively integrate information across multiple documents, a limitation attributed to their constrained reasoning capabilities, particularly in domains such as code generation~\cite{wang2024coderagbench} and in settings involving long-context retrieval~\cite{zhou2025gsm, yen2025helmetevaluatelongcontextlanguage}.

In this work, we tackle the aforementioned challenges by training reasoning language models (RLMs) capable of performing reasoning over a greater number of retrieved documents. Prior approaches for improving RAG have focused on optimizing the retrieval and reranking components by maximizing metrics such as \texttt{recall@5}. In contrast, we propose shifting some of the retrieval burden from the retriever to the generation model itself. An answer generation model that can effectively differentiate between relevant and irrelevant contexts, when given a longer list of retrieved passages, would reduce dependence on high-precision retrieval and increase recall by instead maximizing metrics such as \texttt{recall@10} or \texttt{recall@25}~\cite{jin2025longcontext}. 

Building on the recent success of reinforcement learning (RL) in enhancing the reasoning capabilities of LLMs in the domains of mathematics and coding~\cite{wei2025swerl, xie2025logicrl}, we apply Group Relative Policy Optimization (GRPO;~\citealp{shao2024deepseekmath}) with simple, rule-based rewards to improve the answer generation component of RAG systems. We show that RAG-RL achieves substantial performance gains in both answer and citation generation on three open-domain multi-hop question-answer datasets. Furthermore, our comprehensive evaluation demonstrates that RAG-RL achieves improved performance both in distractor-rich and gold-only settings.\footnote{Gold documents are documents from which the answer to a given question can be deduced, while distractor documents are those that do not contain relevant information.} These settings respectively mirror the use of a weaker retrieval model or a more advanced retrieval and/or reranking systems, and demonstrate that RAG-RL can be used in conjunction with past works on improving retrieval models for further improved performance.

We also conduct a comprehensive study on how different curriculum learning settings affect model performance in post-training. Specifically, we study the effectiveness of introducing question-answer training samples of varying difficulty levels and the impact that the ordering of the training set has on the final performance of the model. We observe that (1) adding easier samples during training teaches the model to more quickly learn how to generate citations as the model no longer has to identify which contexts are relevant, (2) curricula that scale problem difficulty linearly from easiest to hardest perform worse when compared to min-max curricula that begin with the easiest samples and jump straight to the hardest samples, and (3) the benefits of deliberately ordering training samples from easiest to hardest, as proposed by previous curriculum learning studies~\cite{bengio2009curriculumlearning} are not conclusively supported in RL-based post-training. \textit{These empirical observations suggest that constructing training sets of different difficulty levels can increase sample efficiency and generalization by targeting specific skills.}

In summary, the main contributions of this work are as follows:
\begin{itemize}
    \item We introduce RAG-RL, an RLM specifically trained for answer generation in RAG, using RL and curriculum learning.
    \item We benchmark a comprehensive set of different curriculum construction and curriculum learning settings. 
    \item We provide several empirical insights on the effectiveness of different curriculum learning and curriculum construction settings, and how different aspects of the post-training process contribute to final model performance.
\end{itemize}
\section{Related Works}

\subsection{RAG Systems}
Rather than relying solely on parametric knowledge, RAG has been widely used in tasks that require external information~\cite{guu2020realm, lewis2020rag, wang2024rear}. Previous works have made tremendous progress in designing and training sophisticated retrieval and reranking models~\cite{gutiérrez2025ragmemory, weller2025rank1} for open-domain question answering~\cite{chen-etal-2017-reading}. One important line of work has focused on improving the encoder models that are used in the embedding generation process~\cite{lee2025nvembed, muennighoff2025grit}, while another has focused on designing retrieval systems that focus on drawing connections between multiple different documents~\cite{guo2024lightragsimplefastretrievalaugmented, gutiérrez2025ragmemory}. Rank1~\cite{weller2025rank1} has also recently demonstrated that allocating test-time compute for document reranking can lead to performance improvements when retrieving contexts that require in-depth reasoning. Past work has also sought to take advantage of the long context lengths of modern-day LLMs by providing these models with larger sets of retrieved documents, but have shown that these models struggle to effectively identify relevant contexts as the number of retrieved passages increases~\cite{jin2025longcontext, zhou2025gsm, yen2025helmetevaluatelongcontextlanguage}.

\subsection{Multi-Hop Question Answering}
A multi-hop question requires combining information across multiple passages and performing reasoning to arrive at a correct 
answer~\cite{mavi2024mhqa, nishida2019answering}. The number of pieces of information required to successfully answer the question is referred to as the number of hops. The terms passages and documents are used interchangeably to denote disjoint contexts that are retrieved by a retrieval model. Figure~\ref{fig:multihop-floris} demonstrates how RAG-RL operates in the multi-hop question-answering setting.

\subsection{Reasoning Language Models}
With the introduction of RLMs in OpenAI's o1 models~\cite{openai2024openaio1card}, the research community has made progress in replicating similar models that have shown impressive performance in tasks that require reasoning, driven in part due to R1's release~\cite{deepseekai2025deepseekr1incentivizingreasoningcapability}. Prior works have demonstrated the potential for training smaller-scale RLMs in the domains of mathematics, logic, and coding~\cite{xie2025logicrl, wei2025swerl} and have also achieved impressive performance. However, to the best of our knowledge, no one has trained RLMs specifically for the answer generation component of RAG.

\subsection{Curriculum Learning}
Curriculum learning~\cite{bengio2009curriculumlearning} has been extensively studied as a training paradigm that orders training samples by increasing difficulty, leading to improved generalization. In question answering (QA), it has been used to reduce distributional shifts between pre-training and downstream fine-tuning datasets~\cite{zhang2024curriculumdrivendomain}. Recent advances in LLMs have incorporated curriculum-inspired self-improvement mechanisms~\cite{lee2025selfimprovingtransformersovercomeeasytohard}, where models iteratively augment their training data with instances they can already solve, to facilitate generalization to slightly more complex reasoning tasks. In RL, curriculum learning has also been applied to gradually expose agents to more challenging environments~\cite{narvekar2020curriculumlearningreinforcementlearning}; however, its effectiveness remains task-dependent, with some studies reporting only marginal gains~\cite{xie2025logicrl}.
\section{RAG-RL}
In this section, we include a detailed overview of the training process for RAG-RL. We outline the rule-based rewards used in the policy update algorithm and then introduce the curriculum construction settings used in our experiments.

\subsection{Reward Modeling}  

In our work, we use RL as our post-training method as it eliminates the need for training sets consisting of high-quality supervised trajectories produced either by humans or stronger models.

Our rule-based rewards consist of three components: answer rewards, citation rewards, and formatting rewards.

\paragraph{Answer Rewards}
To incentivize correct final answers, we define the answer reward as:
\begin{equation}
\mathcal{R}_{\text{answer}} = \gamma_{\text{answer}} \cdot \mathbb{1}(o_{\text{answer}} = G_{\text{answer}}),
\label{eq:answer_reward}
\end{equation}

where \(o_{\text{answer}}\) is the generated final answer, \(G_{\text{answer}}\) is the ground truth answer, and $\gamma_{answer}$ is a scaling factor, which we set to 5 for our experiments.

\paragraph{Citation Rewards}
To reward correct citations, we define the citation reward as:
\begin{equation}
\resizebox{0.89\columnwidth}{!}{$
\begin{aligned}
\mathcal{R}_{\text{citations}} &= \gamma_{\text{correct}} \cdot 
\text{Recall}(o_{\text{citations}}, G_{\text{citations}}) \\
& - \gamma_{\text{incorrect}} \cdot c_{\text{incorrect}}
\end{aligned}
$}
\label{eq:citation_reward}
\end{equation}

\begin{figure*}[t!]
    \centering
    \includegraphics[trim={0 0 0 0}, clip, scale=0.5]{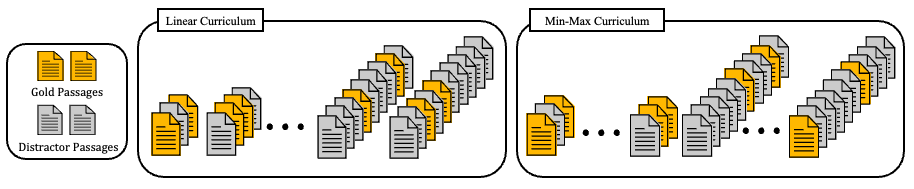}
    \caption{Overview of two curriculum construction settings used during training. Linear denotes a curriculum that scales the difficulty level (the number of distractor passages) from 1 to $K$, while min-max denotes a curriculum that is split evenly between the easiest and the hardest problems.}
    \label{fig:CL}
\end{figure*}

where recall denotes the fraction of relevant citations cited in the final answer $o_{\text{citations}}$, $G_{\text{citations}}$ is the list of ground truth citations, $c_{\text{incorrect}}$ is the number of incorrect citations, and both $\gamma_{\text{correct}}$ and $\gamma_{\text{incorrect}}$ are the scaling factors which we set to 5 and 2 respectively.

\paragraph{Formatting Rewards} 
To enforce the desired output format, we assign a reward of $\gamma_{\text{format}}$ for correct formatting (i.e., the presence of proper XML tags and required headings) while imposing a penalty \( p \) for outputs with excessive text or non-English Unicode characters.\footnote{This penalty has proven particularly beneficial for improving training stability by encouraging the model to generate responses in English.} Formally, we define the reward as:
\begin{equation}
\resizebox{0.89\columnwidth}{!}{$
\mathcal{R}_{\text{formatting}} =
\begin{cases} 
    \gamma_{\text{format}}, & \text{if formatting is correct} \\
    -p, & \text{otherwise.}
\end{cases}
$}
\label{eq:formatting_reward}
\end{equation}

\paragraph{Total Reward and Objective Function}
The overall reward for a training sample is the sum of the individual components:
\[
\mathcal{R}_{\text{total}} = \mathcal{R}_{\text{answer}} + \mathcal{R}_{\text{citation}} + \mathcal{R}_{\text{formatting}}.
\]
This reward is then used in the GRPO algorithm for policy optimization~\cite{shao2024deepseekmath}. The scaling constants we choose in our experiments weigh correctness more than formatting, and in preliminary experiments, we observed no significant changes in performance when adjusting these parameters by small amounts.

\subsection{Curriculum Construction}
\label{curriculum_construction}

Curriculum construction builds training sets by selecting or generating samples across a range of difficulty levels. In this work, we investigate two main difficulty axes: (1) synthetic difficulty, where training samples are algorithmically constructed to span predefined difficulty levels, and (2) accuracy-based difficulty, where the base model’s performance determines which samples it can or cannot solve before any post-training is applied.

\paragraph{Synthetic Difficulty} Given a question $Q$, and a set of documents $D$, we partition $D$ into a set of gold documents $D^+$ and a set of distractor documents $D^-$. The number of hops required to correctly answer $Q$ is given by $j=|D^+|$, while $k=|D^-|$ represents the number of retrieved distractor documents. Naturally, the difficulty of a multi-hop question can be measured along two dimensions: the number of hops required $j$ and the number of distractor documents provided to the generation model from $D^-$. For training RAG-RL, we define the difficulty of a training sample solely based on the subset size of $D^-$ that we provide to the generation model. The easiest training samples contain at most one distractor document along with all gold documents, while the hardest samples include the full set of all retrieved documents. Formally, a training sample $S_i$ of difficulty level $l$ is defined as
\begin{align*}
S_i^l &= [Q, \{D_1^+, D_2^+, \dots, D_j^+, D_1^-, D_2^-, \dots, D_d^-\}], \\
      &\quad \text{where} \quad d = \min\left(\max(l + 2 - j,\, 0),\, k\right),
\end{align*}
where the order of the documents in each $S_i$ is shuffled to ensure a realistic retrieval setting.

Since the minimum number of hops required among all of our datasets is $2$, a difficulty level of $1$ corresponds to 1 distractor document for a 2-hop question. It follows that the highest difficulty level we can effectively introduce is thus $k$, which we denote as $K$ going forward. This definition of synthetic difficulty ensures that all gold contexts are retrieved regardless of difficulty level. 

While the datasets we use contain 2-hop, 3-hop, and 4-hop questions, we focus on the number of distractor documents as the primary axis of difficulty. This choice is motivated by the limited granularity offered by hop count alone, as the vast majority of questions in all three datasets we use are 2-hop questions. For completeness, we include an ablation in Section~\ref{ordering_ablation} that jointly considers both the number of hops and distractor documents by sorting each question by the number of hops and then augmenting each question to span a pre-defined curriculum. Moreover, in Appendix~\ref{app:group-by-num-hops} we present results that show a negative correlation between model performance and the number of hops in each question.

\paragraph{Accuracy-Based Difficulty} An alternative way to define the difficulty of a training sample is to benchmark the base model’s performance on each sample. Specifically, we compute the pass@k of each training sample and partition the dataset into two subsets: samples with pass@k = 1 and samples with pass@k = 0. We refer to the former as \textit{base-answerable} and the latter as \textit{base-unanswerable}. A prediction is considered correct if the generated final answer achieves an F1 score of $1$ when compared to the ground truth answer.
\section{Experiments}

\begin{table*}[t]
    \centering
    \resizebox{\textwidth}{!}{%
    \begin{tabular}{l l c c c c c c c c c c}
        \toprule
        Eval. Setting & Curriculum 
        & \multicolumn{3}{c}{HotpotQA} 
        & \multicolumn{3}{c}{MuSiQue} 
        & \multicolumn{3}{c}{2Wiki} \\
        \cmidrule(lr){3-5} \cmidrule(lr){6-8} \cmidrule(lr){9-11}
        & & Answer F1 & Citation F1 & Joint F1 
        & Answer F1 & Citation F1 & Joint F1 
        & Answer F1 & Citation F1 & Joint F1 \\
        \midrule
        \multirow{6}{*}{\rotatebox{90}{Distractor}} 
        & Baseline & 60.65 & 36.47 & 45.55 & 25.88 & 25.35 & 25.61 & 48.71 & 40.18 & 44.03 \\
        & Max & 66.04 & 73.93 & 69.76 & 40.91 & 53.07 & 46.20 & 67.99 & 77.08 & 72.25 \\
        & Linear & 68.71 & 78.54 & 73.30 & 44.68 & 59.79 & 51.14 & 68.92 & \textbf{83.23} & \textbf{75.40} \\
        & Min-Max & \textbf{68.87} & \textbf{81.64} & \textbf{74.72} & \textbf{47.18} & \textbf{64.48} & \textbf{54.49} & \textbf{70.77} & 76.37 & 73.46 \\
        & Base-Answerable & 66.19 & 72.81 & 69.34 & 41.40 & 54.28 & 46.97 & 68.56 & 76.36 & 72.25 \\
        & Base-Unanswerable & 65.25 & 71.13 & 68.06 & 38.84 & 52.15 & 44.52 & 66.59 & 72.04 & 69.21 \\
        \midrule
        \multirow{6}{*}{\rotatebox{90}{Ideal Retrieval}} 
        & Baseline & 67.90 & 63.26 & 65.50 & 41.16 & 58.16 & 48.21 & 70.29 & 54.49 & 61.39 \\
        & Max & 74.25 & 86.26 & 79.81 & 54.64 & 68.84 & 60.92 & 71.82 & 78.53 & 75.02 \\
        & Linear & 75.67 & 89.34 & 81.94 & 61.10 & 73.90 & 66.89 & 74.31 & \textbf{88.82} & \textbf{80.92} \\
        & Min-Max & \textbf{76.18} & \textbf{93.13} & \textbf{83.81} & \textbf{65.06} & \textbf{81.51} & \textbf{72.37} & \textbf{75.06} & 80.37 & 77.63 \\
        & Base-Answerable & 74.76 & 83.69 & 78.98 & 57.53 & 67.75 & 62.22 & 72.61 & 79.98 & 76.11 \\
        & Base-Unanswerable & 75.23 & 82.99 & 78.92 & 54.53 & 68.37 & 60.67 & 71.04 & 74.30 & 72.63 \\
        \bottomrule
    \end{tabular}%
    }
    \caption{Model performance under the distractor and ideal retrieval evaluation settings across different curriculum construction settings. We use up to 5,000 training samples for all runs as outlined in Section~\ref{training-setup}. The best-performing curriculum for each metric is bolded. Additional training runs with larger training sets are provided in Appendix~\ref{appendix:full_trainsets}.}
    \label{tab:main_results}
\end{table*}

\subsection{Datasets}
We evaluate RAG-RL on three open-domain multi-hop question answering benchmarks: \textbf{HotpotQA}~\cite{yang2018hotpotqa}, \textbf{MuSiQue} (answerable)~\cite{trivedi2022musique}, and \textbf{2Wiki}~\cite{ho-etal-2wiki}. While HotpotQA has been shown to be a weaker test for multi-hop reasoning due to the presence of spurious signals~\cite{trivedi2023ircot}, we include it due to its widespread use but mainly focus on the other two datasets in our discussions.

\subsection{Training Setup} 
\label{training-setup}
We use Qwen2.5-7B-Instruct~\cite{qwen2025qwen25technicalreport} as our base model and employ GRPO for the post-training process. Approximately five thousand QA pairs are sourced from each dataset's respective training sets. All experiments train for a single epoch with a constant learning rate of 1.0e-6, a global batch size of 294, KL coefficient of 0.01, and 7 rollouts for each of the 42 problems in each batch. Each model was trained on Nvidia 8xH100 GPUs. While we observe that reward signals continue to improve beyond 120 steps, we limit training to 5,000 samples due to computational constraints. To assess the generalizability of our experiments to larger training sets, we include additional runs in Appendix~\ref{appendix:full_trainsets}, which confirm that \textit{the trends and insights we report hold consistently at scale}. Additional dataset construction and training details can be found in Appendices~\ref{appendix:datasets} and~\ref{appendix:training}.

\subsection{Baselines}
Our primary baseline is our base model, Qwen2.5-7B-Instruct. For a fine-tuned baseline, we employ the max curriculum, which uses samples at the highest difficulty level $K$, which represents the difficulty of problems expected at test-time. In our tables, we report our base model's performance as ``baseline'' and our fine-tuned baseline as ``max.''

\subsection{Curriculum Learning Settings}
To investigate the effectiveness of curriculum construction in the post-training process, we benchmark several different curricula. As defined in Section~\ref{curriculum_construction}, synthetic difficulty levels range from 1 to $K$, while accuracy-based difficulty partitions training samples into base-answerable and base-unanswerable subsets. Figure~\ref{fig:CL} provides an illustration of the main synthetic curricula used in our experiments. We define a function \( C_{\texttt{setting}}: \{1, \dots, n\} \to \{1, \dots, K\} \) that maps an index \( i \) in the training set to its corresponding difficulty level under each setting. We set $K$ to be 10, 20, and 10 for HotpotQA, MuSiQue, and 2Wiki respectively.

\subsubsection{Synthetic Curricula Variants}
\begin{itemize}
    \item \textbf{Max}: Each sample in the training set is presented at the maximum difficulty level (i.e., the difficulty level expected at test time). Thus, the difficulty function is defined as:
    \[
    C_{\texttt{max}}(i) = K, \quad \forall i \in \{1, \dots, n\}
    \]

    \item \textbf{Linear}: The training set is partitioned into $K$ equally sized subsets, with difficulty levels increasing linearly from \( 1 \) to \( K \). The mapping function is thus:
    \[
    C_{\texttt{linear}}(i) = \left\lceil \frac{K \cdot i}{n} \right\rceil
    \]

    \item \textbf{Min-Max}: The training set is split into two equal parts, where the first half consists of the easiest difficulty level (\( 1 \)) and the second half consists of the hardest difficulty level (\( K \)). The function is defined as:
    \[
    C_{\texttt{min-max}}(i) =
    \begin{cases} 
        1, & \text{if } i \leq n/2 \\
        K, & \text{if } i > n/2
    \end{cases}
    \]
\end{itemize}

\subsubsection{Accuracy-Based Curricula}
\begin{itemize}
    \item \textbf{Base-Answerable}: The training set includes only samples from the maximum synthetic difficulty level that are base-answerable.

    \item \textbf{Base-Unanswerable}: The training set includes only samples from the maximum synthetic difficulty level that are base-unanswerable.
\end{itemize}

\subsection{Evaluation}
To benchmark the performance of our RLMs, we evaluate the F1 scores of the generated answer and passage-level citations on the validation sets provided by our selected benchmark datasets. We sample each response $3$ times and take the average F1 score among all generations. Joint F1, which captures both answer and citation correctness, serves as our primary metric. Dataset statistics can be found in Appendix~\ref{appendix:datasets}.

\paragraph{Comparison to Previous Works} To compare our RLMs to previous works, we measure the performance of our models in two settings: the \textit{distractor setting} and the \textit{ideal retrieval setting}. The distractor setting consists of providing the generation model all gold passages and up to $18$ distractor passages, which is comparable to having the reasoning model handle both reranking and answer generation. On the other hand, in the ideal retrieval setting, the reasoning model is given only the gold truth passages, which is comparable to using a strong retrieval and reranking system. Previous works on improving multi-hop question-answer performance, such as Beam Retrieval~\cite{zhang2024endtoendbeamretrievalmultihop} and Smoothing R3~\cite{yin2023rethinkinglabelsmoothingmultihop}, have primarily focused on optimizing the retrieval component of RAG and utilize span prediction models for answer generation, thus making a direct comparison of generator performance difficult. To better isolate and evaluate generation quality, we adopt the ideal retrieval setting as a more controlled and comparable benchmark. When comparing the performance of RAG-RL to the few past studies that have focused on improving answer generation models~\cite{jin2025longcontext, zhang2024raftadaptinglanguagemodel}, RAG-RL achieves SOTA performance. However, we note that these works may not explicitly focus on multi-hop QA performance nor use the latest base models.
\section{Results}
\label{sec:results}
Our results section is organized into subsections, beginning with the main findings, followed by an investigation of several research questions. The primary goals of our experiments are twofold: first, to evaluate whether curriculum learning can enhance RL-based post-training; and second, to identify which components of the post-training pipeline contribute most significantly to final model performance and why.

\subsection{Main Results}
Table~\ref{tab:main_results} presents the performance of our baseline model and the RLMs we trained under our outlined curriculum construction settings in the distrator and ideal retrieval evaluation settings. The results strongly support the notion that curriculum learning can help improve RL-based post-training. Across all three datasets and both evaluation settings, the min-max and linear curricula achieve the highest joint F1, improving over the max curriculum by a margin of 3 to 8 points.

\begin{figure}[t]
    \centering
    \resizebox{\linewidth}{!}{%
        \includegraphics{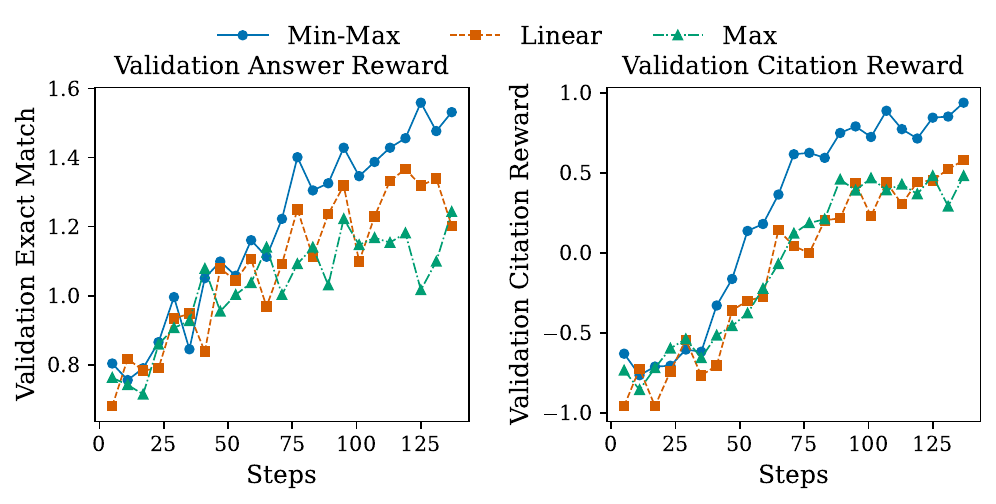}
    }
    \caption{Plots of validation answer and citation rewards during training for three curricula on the MuSiQue dataset.}
    \label{fig:musique_ablation_plots}
\end{figure}

\begin{table}[t]
    \centering
    \resizebox{0.8\linewidth}{!}{%
    \begin{tabular}{l c c c l}
        \toprule
        Metric & Answer F1 & Citation F1 & Joint F1 & \\
        \midrule
        Pass@1 & 25.42 & 25.11 & 25.26 & \\
        Pass@32 & 43.38 & 45.46 & 44.40 & \\
        \bottomrule
    \end{tabular}%
    }
    \caption{Base model pass@k on the MuSiQue dataset.}
    \label{tab:musique_pass32}
\end{table}

\subsection{Does adding easier samples improve performance?}
The primary difference between our synthetic curricula and the max variant is the introduction of easier training samples. The results in Table~\ref{tab:main_results} show that all the curricula that contain easier samples outperform the max curriculum. Figure~\ref{fig:musique_ablation_plots} shows that the validation citation reward climbs much faster for the min-max curriculum, suggesting that training samples with few distractor passages allow the model to learn how to generate correct citations with greater sample efficiency. Given a large number of distractor passages, the model must first identify potential candidates and then reason over them. These findings indicate that easier samples in the synthetic curricula serve to teach citation skills early in training, which can then be refined by harder examples that require multi-step reasoning over longer contexts.

\begin{table}[t]
    \centering
    \resizebox{\linewidth}{!}{%
    \begin{tabular}{l l c c c}
        \toprule
        Eval. Setting & Sample Ordering & Answer F1 & Citation F1 & Joint F1 \\
        \midrule
        \multirow{3}{*}{Distractor}
            & Sorted by Difficulty & 44.68 & 59.79 & 51.14  \\
            & Randomly Shuffled & 44.27 & 60.35 & 51.07 \\
            & Sorted by Hops and Difficulty & 43.78 & 61.20 & 51.04 \\
        \midrule
        \multirow{3}{*}{Ideal}
            & Sorted by Difficulty & 61.10 & 73.90 & 66.89 \\
            & Randomly Shuffled & 60.80 & 73.47 & 66.53 \\
            & Sorted by Hops and Difficulty & 60.94 & 75.16 & 67.31 \\
        \bottomrule
    \end{tabular}%
    }
    \caption{Ablation study on training sample ordering using the linear curriculum on the MuSiQue dataset. Results are reported under both distractor and ideal retrieval settings.}
    \label{tab:musique_ablation}
\end{table}
\subsection{Do we need granular problem difficulty?}
Previous work in the area of self-improvement has shown that LLMs exhibit limited generalizability and that gradually increasing the difficulty levels of training samples from weak-to-strong is effective for helping models generalize beyond their initial training distributions~\cite{lee2025selfimprovingtransformersovercomeeasytohard}. However, our results suggest that this is not always necessary as the min-max curriculum outperforms the linear curriculum in most cases. We believe this to be a byproduct of our base model having relatively strong performance on the task before any post-training is applied. We show our base model's pass@32 in Table~\ref{tab:musique_pass32} and find that these scores are comparable to our fine-tuned baseline.

\subsection{Does dataset ordering matter?}
\label{ordering_ablation}

Curriculum learning strategies where training sets are ordered from easy to hard have been successfully employed in many areas of machine learning~\cite{soviany2022curriculumlearningsurvey}. To assess the impact of sample ordering on model performance, we experiment with three variants of a linear curriculum: (i) samples sorted by difficulty, measured as the number of distractors; (ii) samples presented in a randomly shuffled order; and (iii) samples first ordered by the number of hops and then augmented to span the linear curriculum. As shown in Table~\ref{tab:musique_ablation}, the resulting F1 scores across these curricula are broadly comparable. These findings suggest that, in this setting, the specific ordering of training examples does not yield consistent or significant performance differences.

\subsection{Training on base-answerable versus base-unanswerable samples}
We observe that models trained on base-answerable samples consistently outperform those trained on base-unanswerable samples. As shown in Table~\ref{tab:domain_eval}, this trend holds across both the distractor and ideal retrieval settings. We hypothesize that the superior performance from training on base-answerable samples stems from their alignment with the base model's pretrained capabilities. This is reminiscent of findings from~\citet{zhang2025best}, which demonstrate that supervised fine-tuning is most effective when responses are aligned with the base model’s pretrained distribution. 

\definecolor{custom_orange}{RGB}{209, 83, 23}  
\begin{table*}[t]
    \centering
    \resizebox{\textwidth}{!}{%
    \begin{tabular}{l l c c c c c c c c c c}
        \toprule
        Eval. Setting & Curriculum 
        & \multicolumn{3}{c}{HotpotQA} 
        & \multicolumn{3}{c}{MuSiQue} 
        & \multicolumn{3}{c}{2Wiki} \\
        \cmidrule(lr){3-5} \cmidrule(lr){6-8} \cmidrule(lr){9-11}
        & & Answer F1 & Citation F1 & Joint F1 
        & Answer F1 & Citation F1 & Joint F1 
        & Answer F1 & Citation F1 & Joint F1 \\
        \midrule
        \multirow{6}{*}{\rotatebox{90}{Distractor}}
        & Baseline & {\color{blue}61.31} / {\color{custom_orange}9.21} & {\color{blue}44.89} / {\color{custom_orange}35.65} & {\color{blue}51.83} / {\color{custom_orange}14.64} 
                   & {\color{blue}40.96} / {\color{custom_orange}4.63} & {\color{blue}27.91} / {\color{custom_orange}19.25} & {\color{blue}33.20} / {\color{custom_orange}7.46} 
                   & {\color{blue}61.64} /  {\color{custom_orange}6.75} & {\color{blue}42.71} / {\color{custom_orange}33.80} & {\color{blue}50.45} / {\color{custom_orange}11.25} \\
        & Max & {\color{blue}82.63} /  {\color{custom_orange}18.07} & {\color{blue}76.01} / {\color{custom_orange}68.70} & {\color{blue}79.18} / {\color{custom_orange}28.61} 
              & {\color{blue}63.87} /  {\color{custom_orange}13.51} & {\color{blue}55.54} / {\color{custom_orange}47.51} & {\color{blue}59.41} / {\color{custom_orange}21.04} 
              & {\color{blue}84.42} / {\color{custom_orange}17.37} & {\color{blue}76.53} / {\color{custom_orange}75.50} & {\color{blue}80.28} / {\color{custom_orange}28.25} \\
        & Linear & {\color{blue}86.26} / \textbf{{\color{custom_orange}20.48}} & {\color{blue}80.84} / {\color{custom_orange}73.02} & {\color{blue}83.46} / {\color{custom_orange}31.99} 
                & {\color{blue}70.18} / {\color{custom_orange}18.82} & {\color{blue}62.70} / {\color{custom_orange}53.70} & {\color{blue}66.23} / {\color{custom_orange}27.87} 
                & {\color{blue}85.00} / \textbf{{\color{custom_orange}23.14}} & \textbf{{\color{blue}83.77}} / \textbf{{\color{custom_orange}80.32}} & \textbf{{\color{blue}84.38}} / {\color{custom_orange}\textbf{35.93}} \\
        & Min-Max & \textbf{{\color{blue}86.27}} / {\color{custom_orange}20.40} & \textbf{{\color{blue}83.21}} / \textbf{{\color{custom_orange}76.46}} & \textbf{{\color{blue}84.71}} / \textbf{{\color{custom_orange}32.20}} 
                  & \textbf{{\color{blue}71.73}} / {\color{custom_orange}\textbf{22.89}} & \textbf{{\color{blue}67.42}} / {\color{custom_orange}\textbf{58.47}} & \textbf{{\color{blue}69.51}} / {\color{custom_orange}\textbf{32.90}} 
                  & \textbf{{\color{blue}87.83}} /  {\color{custom_orange}20.41} & {\color{blue}77.66} / {\color{custom_orange}74.19} & {\color{blue}82.43} / {\color{custom_orange}32.02} \\
        & Base-Answerable & {\color{blue}82.63} /  {\color{custom_orange}17.09} & {\color{blue}75.08} / {\color{custom_orange}66.20} & {\color{blue}78.67} / {\color{custom_orange}27.16} 
                    & {\color{blue}65.62} / {\color{custom_orange}16.17} & {\color{blue}57.97} / {\color{custom_orange}47.65} & {\color{blue}61.56} / {\color{custom_orange}24.15}
                    & {\color{blue}86.26} /  {\color{custom_orange}17.42} & {\color{blue}77.29} / {\color{custom_orange}75.14} & {\color{blue}81.53} / {\color{custom_orange}28.28} \\
        & Base-Unanswerable & {\color{blue}82.94} / {\color{custom_orange}16.71} & {\color{blue}72.74} / {\color{custom_orange}65.35} & {\color{blue}77.51} / {\color{custom_orange}26.62} 
                        & {\color{blue}63.59} / {\color{custom_orange}14.49} & {\color{blue}55.44} / {\color{custom_orange}45.81} & {\color{blue}59.23} / {\color{custom_orange}22.01} 
                        & {\color{blue}83.92} /  {\color{custom_orange}16.16} & {\color{blue}71.42} / {\color{custom_orange}73.19} & {\color{blue}77.17} / {\color{custom_orange}26.48} \\
        \bottomrule
    \end{tabular}
    }
    \caption{Model performance under the distractor evaluation setting evaluated on {\color{blue}base-answerable} and {\color{custom_orange}base-unanswerable} evaluation sets. We use up to 5,000 training samples for all runs as outlined in Section~\ref{training-setup}. Results under the ideal retrieval setting can be found in the Appendix in Table~\ref{tab:app_domain_eval}.}
    \label{tab:domain_eval}
\end{table*}

Another possibility for the worse performance from base-unanswerable samples stems from these samples producing a greater number of groups where the rewards among all samples are equal, resulting in a computed advantage of zero and thus no gradient signal nor policy update.
The Dynamic sAmpling Policy Optimization (DAPO;~\citealp{yu2025dapoopensourcellmreinforcement}) algorithm also notes this limitation, and they address this challenge by over-sampling across the entire dataset and discarding prompts that yield uniform rewards across generated responses. Based on our results, we believe tackling this issue from a curriculum construction angle can also yield improved training efficiency.

\begin{table}[t]
    \centering
    \resizebox{\linewidth}{!}{%
    \begin{tabular}{l l c c c}
        \toprule
        Eval. Setting & Curriculum & Answer F1 & Citation F1 & Joint F1 \\
        \midrule
        \multirow{3}{*}{Distractor}
        & Baseline    & 25.88 & 25.35 & 25.61 \\
        & Max         & 40.91 / 37.97   & 53.07 / 36.44   & 46.20 / 37.19 \\
        & Min-Max     & 47.18 / 41.30   & 64.48 / 42.21   & 54.49 / 41.75 \\
        \midrule
        \multirow{3}{*}{Ideal}
        & Baseline    & 41.16 & 58.16 & 48.21 \\
        & Max         & 54.64 / 50.59   & 68.84 / 55.58   & 60.92 / 52.97 \\
        & Min-Max     & 65.06 / 57.95   & 81.51 / 57.18   & 72.37 / 57.56 \\
        \bottomrule
    \end{tabular}%
    }
    \caption{MuSiQue formatting ablation. The first number in each cell denotes the use of both accuracy and formatting rewards, while the second number in each cell denotes the use of only formatting rewards.}
    \label{tab:formatting_ablation}
\end{table}

\subsection{What rule-based rewards matter?}
Recent works have observed that performance gains from using algorithms such as GRPO might stem from improved output formatting rather than improvements in reasoning ability~\cite{petrov2025proofbluffevaluatingllms}. To isolate this phenomenon and see how answer and citation rewards contribute to final model performance, we conduct an ablation where we train our models with only formatting rewards (Eq.~\ref{eq:formatting_reward}), omitting the answer (Eq.~\ref{eq:answer_reward}) and citation rewards (Eq.~\ref{eq:citation_reward}).

Table~\ref{tab:formatting_ablation} presents results on the MuSiQue dataset that demonstrate that employing formatting rewards alone using the min-max curriculum achieves higher answer F1 compared to that of the max curriculum with all rewards. While adding additional answer and citation rewards increases the sample efficiency of the post-training process, this ablation demonstrates that choosing the appropriate training curriculum also plays a major role. We propose that post-training using formatting rewards alone can serve as a stronger baseline before incorporating accuracy-based metrics.

\subsection{Performance on previously unsolved samples}
To evaluate the extent to which RL-based post-training and curriculum learning enhance model performance on previously unsolved questions, we additionally partition the evaluation set into \textit{base-answerable} and \textit{base-unanswerable} subsets. As shown in Table~\ref{tab:domain_eval}, both the min-max and linear curricula yield the greatest gains on the base-unanswerable subset. Nevertheless, a notable performance gap remains between the two subsets. These results suggest that, while curriculum learning can improve a model's ability to address previously challenging examples, generalization to out-of-distribution samples remains a key limitation of RL-based post-training~\cite{xiong2025building}.
\section{Discussion}
In this work, we introduce RAG-RL, a reasoning language model specifically trained for the answer generation component of RAG. Our experiments demonstrate that stronger answer generation models can reduce the burden on retrieval models by reasoning over larger sets of retrieved contexts and that curriculum learning is a powerful tool for improving sample efficiency and generalization during post-training.

Our experiments and ablations support the following key observations: (1) curriculum construction is a powerful method for improving post-training performance, (2) easier training samples (i.e., those with fewer distractor documents) provide models with a stronger signal for learning how to generate citations, and (3) LLMs do not necessarily benefit the most from curricula with gradually increasing difficulty levels (i.e., min-max performs better than a linear curriculum in most of our experiments). While RL-based post-training methods have unlocked a new dimension of scaling for LLMs, our experiments take an in-depth look into understanding which components of these post-training methods contribute to improving model performance and why. 

\paragraph{Future Work}
Our findings suggest that LLMs exhibit limited generalization with performance gains extending only marginally beyond the training distribution. To address this, we propose a systematic categorization of training samples into base-answerable and base-unanswerable instances. Empirically, we find that incorporating synthetic difficulty levels in curriculum construction enables models to acquire citation and reasoning skills with greater sample efficiency and generalizability. Exploring algorithmic curriculum generation methods that target specific areas of improvement, especially for tasks lacking natural difficulty levels, is an exciting research direction.
\section{Limitations}
\label{sec:limitations}
While RAG-RL achieves strong performance across multiple multi-hop QA benchmarks, a key limitation of our experimental setup is the assumption that all relevant gold documents are present in the retrieved set. In real-world scenarios, this assumption may not hold. We suggest explicitly training the model to return an “unanswerable” response when the retrieved context lacks sufficient information. Additionally, our curriculum construction process employs a static progression through difficulty levels, advancing to harder samples regardless of whether the model has fully converged on earlier stages. In contrast, prior work on curriculum learning often adopts adaptive schedulers that revisit easier examples and only proceeds to more difficult ones once the model's performance plateaus. Incorporating such adaptive strategies may further enhance training efficiency and generalization, and we leave this as an avenue for future exploration.

\bibliography{acl_latex}

\newpage
\appendix
\label{sec:appendix}

\section{Dataset Construction}
\label{appendix:datasets}
To construct our synthetic difficulty-based curricula, we randomly sample 5,000 training samples from each of the respective training sets provided by HotpotQA, MuSiQue, and 2Wiki. These datasets each provide distractor passages on similar topics that serve to increase the difficulty of the QA task compared to sampling random passages from their corpora.

To construct the base-answerable and base-unanswerable accuracy-based curricula, we start by randomly selecting up to 40,000 training samples from each of the respective training sets, and we use our base model to partition the questions into those with pass@k = 1 for base-answerable and pass@k = 0 for base-unanswerable. We take the maximum answer F1 score across 8 generations for each question to assess pass@k. We then clip the number of training samples so that both the base-answer and base-unanswerable training sets have the same number of training samples (approximately 5,000).

We provide the dataset statistics for all of our datasets in Tables~\ref{app:training_sets} and~\ref{app:evaluation_sets} and note that the respective number of samples in each dataset are dependent on the base model's performance, as well as the respective sizes of each dataset's original training and evaluation set sizes.

\begin{table}[htbp]
    \centering
    \resizebox{\linewidth}{!}{
    \begin{tabular}{lccc}
        \toprule
        & \multicolumn{3}{c}{Number of Training Samples} \\
        \cmidrule(lr){2-4}
        Curriculum & HotpotQA & MuSiQue & 2Wiki \\
        \midrule
        Max      & 5000 & 5000 & 5000 \\
        Linear   & 5000 & 5000 & 5000 \\
        Min-Max  & 5000 & 5000 & 5000 \\
        Base-Answerable       & 3991 & 4057 & 4468 \\
        Base-Unanswerable      & 3991 & 4057 & 4468 \\
        \bottomrule
    \end{tabular}
    }
    \caption{Number of training samples for each curriculum and dataset.}
    \label{app:training_sets}
\end{table}

\begin{table}[htbp]
    \centering
    \resizebox{\linewidth}{!}{
    \begin{tabular}{lccc}
        \toprule
        & \multicolumn{3}{c}{Number of Evaluation Samples} \\
        \cmidrule(lr){2-4}
        Setting & HotpotQA & MuSiQue & 2Wiki \\
        \midrule
        Distractor               & 1000 & 1000 & 1000 \\
        Ideal Retrieval          & 1000 & 1000 & 1000 \\
        \makecell[l]{Distractor \\ (Base-Answerable/Unanswerable)}      & 810  & 681  & 1000 \\
        \makecell[l]{Ideal Retrieval \\ (Base-Answerable/Unanswerable)} & 512  & 379  & 1000 \\
        \bottomrule
    \end{tabular}
    }
    \caption{Number of evaluation samples for each dataset and evaluation setting.}
    \label{app:evaluation_sets}
\end{table}

\section{Training}
\label{appendix:training}
We use the Axolotl package\footnote{\url{https://github.com/axolotl-ai-cloud/axolotl}} to conduct our post-training. For GRPO specifically, we use a beta of 0.01 and a clippling parameter of 0.2. A full list of all the hyperparameters used during training can be found in our code in the supplementary materials. We plan to release all of our code, data, and models as an open-source GitHub repository at the conclusion of the review process.

\section{Additional Experiments}
\subsection{Training on Larger Train Sets}
\label{appendix:full_trainsets}
The training runs presented in Section~\ref{sec:results} are based on subsets of up to 5,000 samples, constrained by computational limitations and the large number of post-training runs required across our curricula. To assess the scalability of our findings to larger training set sizes, we include additional results in Table~\ref{tab:appendix_main_results}, using 40,000 training samples for HotpotQA and 19,900 for MuSiQue. These results confirm that the trends and conclusions in the main paper continue to hold at larger scales. We do not include experiments with base-answerable and base-unanswerable subsets in this setting due to the reduced number of samples remaining after filtering for base-unanswerable instances.

\subsection{Performance on Base-Answerable and Base-Unanswerable Questions}
Table~\ref{tab:appendix_main_results} reports the complete results for both the distractor and ideal retrieval settings, evaluated separately on base-answerable and base-unanswerable questions. Notably, in the ideal retrieval setting, answer F1 scores are lower than those observed in the distractor setting. Upon closer analysis, we attribute this trend to limitations of the F1 metric, which can assign partial credit to answers that are incorrect. When we replaced F1 with the exact match (EM) metric, we observed smaller performance gains on the base-unanswerable subsets and more comparable results across the two retrieval settings.

\subsection{Model Performance Grouped by Number of Hops}
\label{app:group-by-num-hops}
Table~\ref{tab:app_performance_musique} includes the results from Appendix~\ref{appendix:full_trainsets} when grouping by the number of hops required by each question. The results show that as the number of hops increases per question, model performance consistently decreases. However, the models trained with the min-max curriculum still achieve the highest F1 scores.

\begin{table*}[t]
    \centering
    \resizebox{\textwidth}{!}{%
    \begin{tabular}{l l c c c c c c}
        \toprule
        Setting & Curriculum 
        & \multicolumn{3}{c}{HotpotQA} 
        & \multicolumn{3}{c}{MuSiQue} \\
        \cmidrule(lr){3-5} \cmidrule(lr){6-8}
        & & Answer F1 & Citation F1 & Joint F1 
          & Answer F1 & Citation F1 & Joint F1 \\
        \midrule
        \multirow{4}{*}{\rotatebox{90}{Distractor}} 
        & Baseline & 60.65 & 36.47 & 45.55 & 25.88 & 25.35 & 25.61 \\
        & Max & 68.52 & 71.55 & 70.00 & 46.06 & 64.66 & 53.80 \\
        & Linear & 72.65 & 80.53 & 76.39 & 47.93 & 68.45 & 56.38 \\
        & Min-Max & \textbf{74.97} & \textbf{81.25} & \textbf{77.98} & \textbf{55.13} & \textbf{69.27} & \textbf{61.40} \\
        \midrule
        \multirow{4}{*}{\rotatebox{90}{Ideal Ret.}} 
        & Baseline & 67.90 & 63.26 & 65.50 & 41.16 & 58.16 & 48.21 \\
        & Max & 74.79 & 77.38 & 76.06 & 59.04 & 77.99 & 67.21 \\
        & Linear & 77.94 & 86.45 & 81.97 & 64.84 & 85.23 & 73.65 \\
        & Min-Max & \textbf{79.74} & \textbf{87.38} & \textbf{83.38} & \textbf{69.79} & \textbf{86.81} & \textbf{77.37} \\
        \bottomrule
    \end{tabular}%
    }
    \caption{Model performance under the distractor and ideal retrieval settings for different curriculum construction settings given larger training set sizes. We use between 20,000-40,000 training samples for all runs as outlined in Appendix~\ref{appendix:full_trainsets}.}
    \label{tab:appendix_main_results}
\end{table*}

\section{Example Prompts and Outputs}
\label{appendix:prompts_and_outputs}

The system prompt and user instructions we use to instruct our models are included in Figures~\ref{fig:system-instructions} and~\ref{fig:user-instructions}.

\begin{figure}[htbp]
  \centering
  \begin{minipage}{\columnwidth}
    \begin{tcolorbox}[colback=violet!5, colframe=blue!75!black, title=\textbf{\textcolor{white}{System Prompt}}, coltitle=white, fonttitle=\bfseries]
    \footnotesize
    Respond in the following format:\\
    <reasoning>\\
    ...\\
    </reasoning>\\
    <answer>\\
    Final answer: final answer\\
    Supporting passages: title1, title2,...\\
    </answer>
    \end{tcolorbox}
  \end{minipage}
  \caption{System prompt used for all experiments.}
  \label{fig:system-instructions}
\end{figure}

\begin{figure}[htbp]
  \centering
  \begin{minipage}{\columnwidth}
    \begin{tcolorbox}[colback=violet!5, colframe=blue!75!black, title=\textbf{\textcolor{white}{User Instructions}}, coltitle=white, fonttitle=\bfseries]
    \footnotesize
    Answer the question using only the provided passages. Verify your answer directly against the text, and cite only the passages you used in your final answer.
    \end{tcolorbox}
  \end{minipage}
  \caption{User instructions used for all experiments.}
  \label{fig:user-instructions}
\end{figure}

\definecolor{custom_orange}{RGB}{209, 83, 23}  
\begin{table*}[t!]
    \centering
    \resizebox{\textwidth}{!}{
    \begin{tabular}{l l c c c c c c c c c c}
        \toprule
        Eval. Setting & Curriculum 
        & \multicolumn{3}{c}{HotpotQA} 
        & \multicolumn{3}{c}{MuSiQue} 
        & \multicolumn{3}{c}{2Wiki} \\
        \cmidrule(lr){3-5} \cmidrule(lr){6-8} \cmidrule(lr){9-11}
        & & Answer F1 & Citation F1 & Joint F1 
        & Answer F1 & Citation F1 & Joint F1 
        & Answer F1 & Citation F1 & Joint F1 \\
        \midrule
        \multirow{6}{*}{\rotatebox{90}{Distractor}}
        & Baseline & {\color{blue}61.31} / {\color{custom_orange}9.21} & {\color{blue}44.89} / {\color{custom_orange}35.65} & {\color{blue}51.83} / {\color{custom_orange}14.64} 
                   & {\color{blue}40.96} / {\color{custom_orange}4.63} & {\color{blue}27.91} / {\color{custom_orange}19.25} & {\color{blue}33.20} / {\color{custom_orange}7.46} 
                   & {\color{blue}61.64} /  {\color{custom_orange}6.75} & {\color{blue}42.71} / {\color{custom_orange}33.80} & {\color{blue}50.45} / {\color{custom_orange}11.25} \\
        & Max & {\color{blue}82.63} /  {\color{custom_orange}18.07} & {\color{blue}76.01} / {\color{custom_orange}68.70} & {\color{blue}79.18} / {\color{custom_orange}28.61} 
              & {\color{blue}63.87} /  {\color{custom_orange}13.51} & {\color{blue}55.54} / {\color{custom_orange}47.51} & {\color{blue}59.41} / {\color{custom_orange}21.04} 
              & {\color{blue}84.42} / {\color{custom_orange}17.37} & {\color{blue}76.53} / {\color{custom_orange}75.50} & {\color{blue}80.28} / {\color{custom_orange}28.25} \\
        & Linear & {\color{blue}86.26} / \textbf{{\color{custom_orange}20.48}} & {\color{blue}80.84} / {\color{custom_orange}73.02} & {\color{blue}83.46} / {\color{custom_orange}31.99} 
                & {\color{blue}70.18} / {\color{custom_orange}18.82} & {\color{blue}62.70} / {\color{custom_orange}53.70} & {\color{blue}66.23} / {\color{custom_orange}27.87} 
                & {\color{blue}85.00} / \textbf{{\color{custom_orange}23.14}} & \textbf{{\color{blue}83.77}} / \textbf{{\color{custom_orange}80.32}} & \textbf{{\color{blue}84.38}} / {\color{custom_orange}\textbf{35.93}} \\
        & Min-Max & \textbf{{\color{blue}86.27}} / {\color{custom_orange}20.40} & \textbf{{\color{blue}83.21}} / \textbf{{\color{custom_orange}76.46}} & \textbf{{\color{blue}84.71}} / \textbf{{\color{custom_orange}32.20}} 
                  & \textbf{{\color{blue}71.73}} / {\color{custom_orange}\textbf{22.89}} & \textbf{{\color{blue}67.42}} / {\color{custom_orange}\textbf{58.47}} & \textbf{{\color{blue}69.51}} / {\color{custom_orange}\textbf{32.90}} 
                  & \textbf{{\color{blue}87.83}} /  {\color{custom_orange}20.41} & {\color{blue}77.66} / {\color{custom_orange}74.19} & {\color{blue}82.43} / {\color{custom_orange}32.02} \\
        & Base-Answerable & {\color{blue}82.63} /  {\color{custom_orange}17.09} & {\color{blue}75.08} / {\color{custom_orange}66.20} & {\color{blue}78.67} / {\color{custom_orange}27.16} 
                    & {\color{blue}65.62} / {\color{custom_orange}16.17} & {\color{blue}57.97} / {\color{custom_orange}47.65} & {\color{blue}61.56} / {\color{custom_orange}24.15}
                    & {\color{blue}86.26} /  {\color{custom_orange}17.42} & {\color{blue}77.29} / {\color{custom_orange}75.14} & {\color{blue}81.53} / {\color{custom_orange}28.28} \\
        & Base-Unanswerable & {\color{blue}82.94} / {\color{custom_orange}16.71} & {\color{blue}72.74} / {\color{custom_orange}65.35} & {\color{blue}77.51} / {\color{custom_orange}26.62} 
                        & {\color{blue}63.59} / {\color{custom_orange}14.49} & {\color{blue}55.44} / {\color{custom_orange}45.81} & {\color{blue}59.23} / {\color{custom_orange}22.01} 
                        & {\color{blue}83.92} /  {\color{custom_orange}16.16} & {\color{blue}71.42} / {\color{custom_orange}73.19} & {\color{blue}77.17} / {\color{custom_orange}26.48} \\
        \midrule
        \multirow{6}{*}{\rotatebox{90}{Ideal Retrieval}} 
        & Baseline & {\color{blue}75.17} / {\color{custom_orange}5.15} & {\color{blue}66.03} / {\color{custom_orange}64.70} & {\color{blue}70.30} / {\color{custom_orange}9.55} 
                   & {\color{blue}60.77} / {\color{custom_orange}4.66} & {\color{blue}52.22} / {\color{custom_orange}52.02} & {\color{blue}56.17} / {\color{custom_orange}8.55} 
                   & {\color{blue}70.29} /  {\color{custom_orange}6.23} & {\color{blue}54.49} / {\color{custom_orange}54.49} & {\color{blue}61.39} / {\color{custom_orange}11.19} \\
        & Max & {\color{blue}89.74} / {\color{custom_orange}8.61} & {\color{blue}88.51} / {\color{custom_orange}86.67} & {\color{blue}89.12} / {\color{custom_orange}15.67} 
              & {\color{blue}77.67} / {\color{custom_orange}12.14} & {\color{blue}69.11} / {\color{custom_orange}70.15} & {\color{blue}73.14} / {\color{custom_orange}20.69} 
              & {\color{blue}88.64} / {\color{custom_orange}13.56} & {\color{blue}79.84} / {\color{custom_orange}84.54} & {\color{blue}84.01} / {\color{custom_orange}23.36} \\
        & Linear & {\color{blue}90.27} / \textbf{{\color{custom_orange}10.66}} & {\color{blue}91.13} / {\color{custom_orange}89.32} & {\color{blue}90.70} / \textbf{{\color{custom_orange}19.04}} 
                & {\color{blue}81.70} / {\color{custom_orange}19.65} & {\color{blue}74.24} / {\color{custom_orange}75.10} & {\color{blue}77.79} / {\color{custom_orange}31.14} 
                & {\color{blue}90.50} / \textbf{\color{custom_orange}19.70} & \textbf{{\color{blue}88.34}} / \textbf{{\color{custom_orange}91.37}} & \textbf{{\color{blue}89.41}} / {\color{custom_orange}\textbf{32.41}} \\
        & Min-Max & {\color{blue}\textbf{92.06}} / {\color{custom_orange}10.30} & \textbf{{\color{blue}95.12}} / {\color{custom_orange}\textbf{93.68}} & \textbf{{\color{blue}93.56}} / {\color{custom_orange}18.56} 
                  & {\color{blue}\textbf{86.46}} / {\color{custom_orange}\textbf{29.55}} & \textbf{{\color{blue}\textbf{81.60}}} / {\color{custom_orange}\textbf{81.64}} & {\color{blue}\textbf{83.96}} / {\color{custom_orange}\textbf{43.39}} 
                  & {\color{blue}\textbf{91.01}} /  {\color{custom_orange}17.17} & {\color{blue}80.95} / {\color{custom_orange}83.03} & {\color{blue}85.69} / {\color{custom_orange}28.46} \\
        & Base-Answerable & {\color{blue}89.74} /  {\color{custom_orange}8.87} & {\color{blue}85.27} / {\color{custom_orange}83.83} & {\color{blue}87.45} / {\color{custom_orange}16.04} 
                    & {\color{blue}79.11} / {\color{custom_orange}14.17} & {\color{blue}67.38} / {\color{custom_orange}67.07} & {\color{blue}72.77} / {\color{custom_orange}23.39} 
                    & {\color{blue}89.68} /  {\color{custom_orange}13.76} & {\color{blue}80.61} / {\color{custom_orange}85.46} & {\color{blue}84.90} / {\color{custom_orange}23.70} \\
        & Base-Unanswerable & {\color{blue}88.89} /  {\color{custom_orange}7.77} & {\color{blue}84.98} / {\color{custom_orange}83.93} & {\color{blue}86.89} / {\color{custom_orange}14.22} 
                        & {\color{blue}77.66} / {\color{custom_orange}12.12} & {\color{blue}68.00} / {\color{custom_orange}70.96} & {\color{blue}72.51} / {\color{custom_orange}20.71} 
                        & {\color{blue}88.95} /  {\color{custom_orange}11.88} & {\color{blue}75.07} / {\color{custom_orange}81.82} & {\color{blue}81.42} / {\color{custom_orange}20.75} \\
        \bottomrule
    \end{tabular}%
    }
    \caption{Model performance under the distractor and ideal retrieval evaluation settings evaluated on {\color{blue}base-answerable} and {\color{custom_orange}base-unanswerable} evaluation sets. We use up to 5,000 training samples for all runs as outlined in Section~\ref{training-setup}} 
    \label{tab:app_domain_eval}
\end{table*}

\begin{table*}[t]
    \centering
    \resizebox{\textwidth}{!}{%
    \begin{tabular}{l c c c c c c c c c}
        \toprule
         & \multicolumn{3}{c}{MuSiQue 2-hop} & \multicolumn{3}{c}{MuSiQue 3-hop} & \multicolumn{3}{c}{MuSiQue 4-hop} \\
        \cmidrule(lr){2-4} \cmidrule(lr){5-7} \cmidrule(lr){8-10}
        Curriculum & Answer F1 & Citation F1 & Joint F1 & Answer F1 & Citation F1 & Joint F1 & Answer F1 & Citation F1 & Joint F1 \\
        \midrule
        Baseline & 28.94 & 29.46 & 29.19 & 23.09 & 22.62 & 22.85 & 21.65 & 17.80 & 19.53 \\
        Max & 48.95 & 70.47 & 57.77 & 44.94 & 63.84 & 52.75 & 39.22 & 48.23 & 43.26 \\
        Linear & 52.04 & 74.38 & 61.23 & 45.53 & 67.08 & 54.24 & 39.74 & \textbf{52.74} & 45.33 \\
        Min-Max & \textbf{57.03} & \textbf{76.27} & \textbf{65.26} & \textbf{54.11} & \textbf{67.51} & \textbf{60.08} & \textbf{51.16} & 50.94 & \textbf{51.05} \\
        \bottomrule
    \end{tabular}%
    }
    \caption{Model performance on MuSiQue in the \textit{distractor setting} grouped by the number of hops in each question. We use between 20,000-40,000 training samples for all the runs in this table.}
    \label{tab:app_performance_musique}
\end{table*}

\end{document}